\documentclass[nonacm,sigconf,screen,table]{acmart}

\AtBeginDocument{%
  }
    
\usepackage{multirow} 
\usepackage{tabularx}
\usepackage{booktabs}
\renewcommand\footnotetextcopyrightpermission[1]{}

\begin{document}

\title{Keep the General, Inject the Specific: Structured Dialogue Fine-Tuning for Knowledge Injection without Catastrophic Forgetting}

\author{Yijie Hong}
\authornote{Both authors contributed equally to this research.}
\email{1656125037@sjtu.edu.cn}
\affiliation{%
  \institution{Shanghai Jiao Tong University}
  \city{Shanghai}
  \country{China}
}
\author{Xiaofei Yin}
\authornotemark[1]
\email{yinxiaofei.yxf@antgroup.com}
\affiliation{%
  \institution{Ant Security Lab, Ant Group}
  \city{Shanghai}
  \country{China}
}
\author{Xinzhong Wang}
\email{2046449167@sjtu.edu.cn}
\affiliation{%
  \institution{Shanghai Jiao Tong University}
  \city{Shanghai}
  \country{China}
}
\author{Yi Tu}
\email{qianyi.ty@antgroup.com}
\affiliation{%
  \institution{Ant Security Lab, Ant Group}
  \city{Shanghai}
  \country{China}
}
\author{Ya Guo}
\email{guoya.gy@antgroup.com}
\affiliation{%
  \institution{Ant Security Lab, Ant Group}
  \city{Shanghai}
  \country{China}
}
\author{Weiqiang Wang}
\email{weiqiang.wwq@antgroup.com}
\affiliation{%
  \institution{Ant Security Lab, Ant Group}
  \city{Shanghai}
  \country{China}
}
\author{Sufeng Duan}
\email{1140339019dsf@sjtu.edu.cn}
\affiliation{%
  \institution{Shanghai Jiao Tong University}
  \city{Shanghai}
  \country{China}
}
\author{Lingyong Fang}
\email{fangly@sjtu.edu.cn}
\affiliation{%
  \institution{Shanghai Jiao Tong University}
  \city{Shanghai}
  \country{China}
}
\author{Depeng Wang}
\email{wdp432379@antgroup.com}
\affiliation{%
  \institution{Ant Security Lab, Ant Group}
  \city{Shanghai}
  \country{China}
}
\author{Huijia Zhu}
\email{huijia.zhj@antgroup.com}
\authornotemark[2]
\affiliation{%
  \institution{Ant Security Lab, Ant Group}
  \city{Shanghai}
  \country{China}
}

\renewcommand{\shortauthors}{Hong et al.}

\begin{abstract}
Large Vision Language Models have demonstrated impressive versatile capabilities through extensive multimodal pre-training, but face significant limitations when incorporating specialized knowledge domains beyond their training distribution. These models struggle with a fundamental dilemma: direct adaptation approaches that inject domain-specific knowledge often trigger catastrophic forgetting of foundational visual-linguistic abilities. We introduce \textbf{S}tructured \textbf{D}ialogue \textbf{F}ine-\textbf{T}uning (SDFT), an effective approach that effectively injects domain-specific knowledge while minimizing catastrophic forgetting. Drawing inspiration from supervised fine-tuning in LLMs and subject-driven personalization in text-to-image diffusion models, our method employs a three-phase dialogue structure: Foundation Preservation reinforces pre-trained visual-linguistic alignment through caption tasks; Contrastive Disambiguation introduces carefully designed counterfactual examples to maintain semantic boundaries; and Knowledge Specialization embeds specialized information through chain-of-thought reasoning. Experimental results across multiple domains confirm SDFT's effectiveness in balancing specialized knowledge acquisition with general capability retention. Our key contributions include a data-centric dialogue template that balances foundational alignment with targeted knowledge integration, a weighted multi-turn supervision framework, and comprehensive evaluation across diverse knowledge types.
\end{abstract}




\maketitle

\section{Introduction}
\begin{figure}[t]
  \centering
  \includegraphics[width=\linewidth]{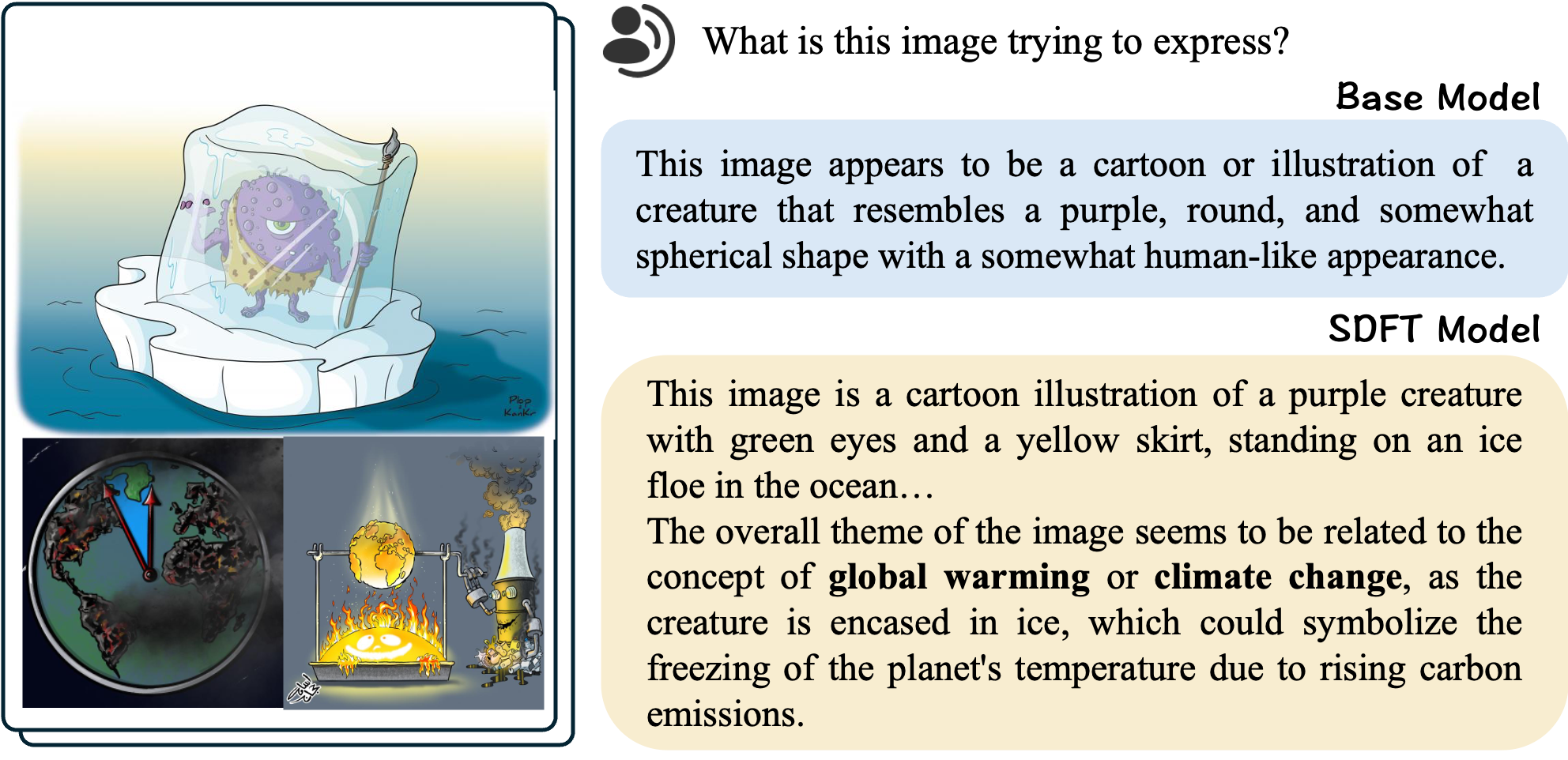}
\caption{Structured multi-turn supervision enables knowledge injection without forgetting. The base LVLM (Qwen2-VL-2B) describes only surface-level content, failing to capture the deeper conceptual meaning (e.g., global warming). In contrast, the same model fine-tuned with our SDFT approach identifies the symbolic implications by linking visual elements to abstract concepts. }
  \label{fig:first_v4}
\end{figure}
Recent advances in Large Vision-Language Models (LVLMs) have demonstrated remarkable capabilities across general-purpose visual understanding tasks \cite{liu2024llava,bai2023qwen}. These models excel at recognizing common objects, describing scenes, and answering straightforward questions about visual content. Their impressive performance stems from extensive pre-training on diverse multimodal datasets that capture broad patterns of visual-linguistic correspondence.
Despite these achievements, LVLMs face inherent limitations imposed by their pre-training distribution. Like text-based language models, they are constrained by the scope and diversity of their training data. The multimodal corpora that form the foundation of LVLM pre-training are limited snapshots of general knowledge, lacking depth in specialized domains and expert knowledge areas.

The conventional approach to addressing these knowledge limitations is through fine-tuning, which adapts pre-trained models to specialized domains using task-specific data. While fine-tuning can inject target knowledge, it frequently triggers catastrophic forgetting—a phenomenon where the model's newly acquired capabilities come at the expense of its foundational abilities \cite{Zhou_2025, Kirkpatrick_2017}. This degradation of general performance represents a fundamental dilemma in knowledge injection. Furthermore, training separate specialized models for each knowledge domain is computationally inefficient, particularly when the target knowledge is relatively limited in scope. Alternative approaches such as retrieval-augmented generation (RAG) \cite{gao2024retrievalaugmentedgenerationlargelanguage} introduce operational overhead, struggle with noisy retrievals, and cannot effectively handle fine-grained visual distinctions without extensive annotated databases \cite{xia2025mmedragversatilemultimodalrag, he2025analyzingboostingpowerfinegrained}. This challenge necessitates a novel approach that can effectively inject specialized knowledge while preserving general capabilities—essentially, a method that allows us to "keep the general, inject the specific." The ideal solution would enable LVLMs to acquire domain-specific expertise without compromising their foundational visual-linguistic intelligence, creating more versatile and adaptable systems for practical applications.

We propose Structured Dialogue Fine-Tuning (SDFT), a data-centric approach that resolves the catastrophic forgetting dilemma through carefully designed structured dialogues. Our approach draws inspiration from personalization techniques in text-to-image diffusion models, particularly DreamBooth \cite{ruiz2023dreamboothfinetuningtexttoimage}, which binds specific visual concepts to unique identifiers (e.g., "a [V] dog") while preserving the model's general knowledge about common concepts (e.g., "a dog"). This binding mechanism prevents knowledge contamination and maintains semantic boundaries between specialized and general knowledge.

Our key insight is that controlled exposure to complementary knowledge during fine-tuning serves as an effective regularizer, enabling the model to distinguish domain-invariant patterns from specialized knowledge. As illustrated in Figure~\ref{fig:first_v4}, we design a three-phase structured dialogue template that mimics this knowledge isolation strategy: (1) \textbf{Foundation Preservation} reinforces the model's pre-trained visual-linguistic alignment through caption tasks; (2) \textbf{Contrastive Disambiguation} introduces carefully designed counterfactual examples where target knowledge is replaced with unrelated ones (e.g., "Transportation"), creating valuable negative samples; and (3) \textbf{Knowledge Specialization} introduces high-fidelity question-answer pairs that embed the specialized information (e.g., "Global Warming") with chain-of-thought reasoning.

To comprehensively evaluate our approach, we examine three distinct knowledge injection scenarios that represent progressively complex challenges in visual understanding:
First, we address \textbf{personalized entity recognition}, where models must identify specific instances (e.g., "my pet cat Max") while maintaining general object understanding \cite{yin2025insightvisioncomprehensivemultilevelchinesebased}. This represents the foundation of knowledge injection—teaching models to recognize specific entities without compromising their general categorization abilities.
Second, we tackle \textbf{abstract concept understanding}, where models must connect visual elements to symbolic meanings \cite{ling2024domainspecializationkeymake}. This more challenging task requires models to bridge perceptual features with conceptual interpretations, such as recognizing that images of factory emissions represent environmental concerns beyond their visible elements.
Third, we explore \textbf{domain expertise integration} in biomedical contexts, where specialized terminology and complex reasoning patterns are required \cite{song2025injectingdomainspecificknowledgelarge}. This represents the most advanced form of knowledge injection, demanding the integration of professional expertise for accurate visual interpretation.

\noindent\textbf{Our key contributions are as follows:}
\begin{list}{$\bullet$}{\setlength{\leftmargin}{1em}
                       \setlength{\itemindent}{0em}
                       \setlength{\labelsep}{0.5em}
                       \setlength{\parsep}{0pt}
                       \setlength{\itemsep}{0pt}}
\item A novel data-centric fine-tuning strategy that effectively injects specialized knowledge into LVLMs while minimizing catastrophic forgetting.
\item The introduction of a structured dialogue template balancing foundational visual-linguistic alignment with targeted knowledge integration through controlled knowledge disambiguation.
\item Development of a weighted multi-turn supervision framework preserving general capabilities throughout the specialization process.
\item  Comprehensive experimental validation across diverse knowledge types, demonstrating the versatility and effectiveness of our approach in balancing specialized knowledge acquisition with general capability retention.
\end{list}

\section{Related Work}
\label{sec:related_work}
\subsection{Text-to-Image Personalization}
Personalization in image generation aims to incorporate personalized concept into pre-trained text-to-image diffusion models to generate specific personalized concept in various contexts. Methods for personalized text-to-image generation have been widely explored. Early approaches like Textual Inversion and Dreambooth\cite{gal2022imageworthwordpersonalizing, ruiz2023dreamboothfinetuningtexttoimage} require training for each personalized concept, leading to scalability issues. To avoid test-time fine-tuning, some methods \cite{shi2023instantboothpersonalizedtexttoimagegeneration, ye2023ipadaptertextcompatibleimage, gal2023encoderbaseddomaintuningfast} use pre-trained vision encoders to encode personalized concepts, integrating the encoded features into diffusion model components through word embeddings or network parameters to facilitate the generation of personalized content. Other methods \cite{shi2023instantboothpersonalizedtexttoimagegeneration, zeng2024jedijointimagediffusionmodels, he2024imagineyourselftuningfreepersonalized} avoid test-time fine-tuning through personalized pre-training. Similarly, our proposed approach for personalizing VLMs can avoid test-time fine-tuning and effectively address scalability issues.
\subsection{Personalized Large Vision Language Models}
Personalization in LVLMs aims to develop models capable of distinguishing specific visual identities without explicit prompts. MyVLM \cite{alaluf2024myvlmpersonalizingvlmsuserspecific} introduces a concept head over CLIP tokens to represent user-specific entities, but requires test-time fine-tuning for adaptation. Similarly, Yo’LLaVA \cite{nguyen2024yollavapersonalizedlanguagevision} augments token embeddings to encode personalized object descriptions. Both approaches rely on textual inversion-like techniques \cite{gal2022imageworthwordpersonalizing}, which constrain scalability by supporting only one concept per training session and requiring test-time updates. RAP \cite{hao2025rapretrievalaugmentedpersonalizationmultimodal} mitigates this by removing test-time training through large-scale pretraining, but its reliance on nearest reference matching in CLIP space can hinder robust contextual understanding across images. While encoder-based methods like PVLM \cite{pi2024personalizedvisualinstructiontuning} improve efficiency by leveraging frozen encoders, they remain limited in capturing fine-grained personalization without significant supervision.In contrast, our method enables concept-level adaptation through multi-turn dialogue supervision without requiring test-time tuning or retrieval modules. It generalizes across multiple personalized concepts while preserving general vision-language capabilities, addressing the scalability and contextuality challenges of prior methods.

\subsection{Knowledge Injection in Language Models}
Recent work on knowledge injection in LLMs has explored post-training strategies to enhance factual accuracy. These include continued pretraining with knowledge-infilling objectives \cite{xu2023kilm}, factuality aware preference optimization  \cite{tian2023fine, rafailov2023direct}, and unsupervised absorption of paraphrased, post-cutoff corpora \cite{ovadia2023fine}. While effective in textual domains, these approaches primarily focus on language-only settings and do not address the challenges of multimodal alignment in vision-language models.
In contrast to LLMs, knowledge injection in LVLMs remains underexplored.AdaMLLM \cite{cheng2025domainspecificposttrainingmultimodallarge} represents an early attempt to adapt LVLMs to domain-specific tasks via two-round dialogues combining general and specialized data. However, it primarily focuses on domain adaptation rather than explicit knowledge isolation, and lacks mechanisms to preserve general capabilities. RAG \cite{xia2025mmedragversatilemultimodalrag}offers another strategy by dynamically incorporating external information during inference, but it introduces latency and struggles with fine-grained visual grounding, particularly when retrieval results are noisy or incomplete.These limitations highlight the need for a unified knowledge injection framework that enables LVLMs to acquire new concepts while retaining their general vision-language grounding. To this end, we propose SDFT that injects domain-specific knowledge through multi-turn supervision, explicitly balancing specialization and retention.
\begin{figure*}[t]
  \centering
  \includegraphics[width=0.9\linewidth]{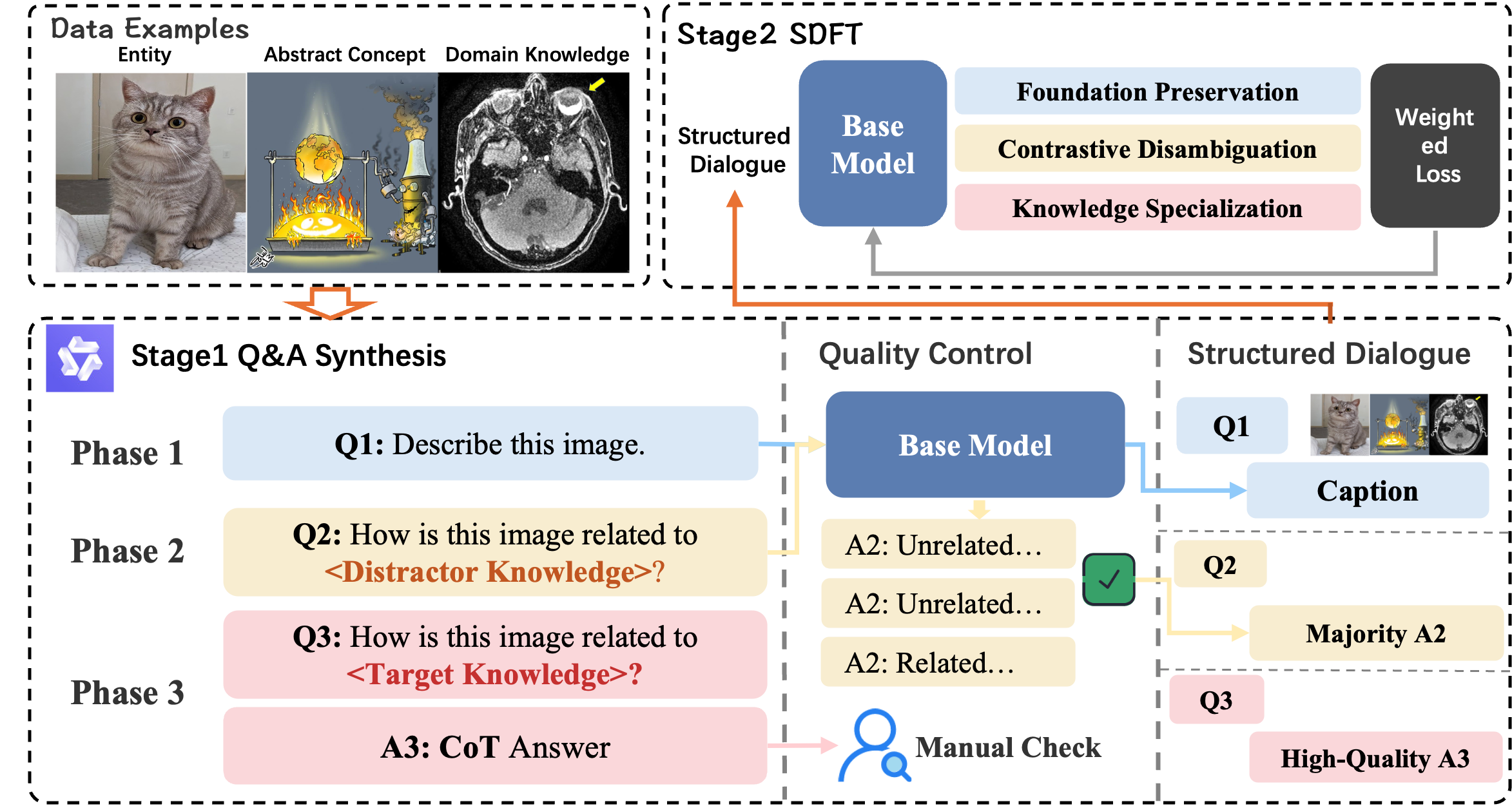}
  \caption{Overview of the SDFT framework. Given domain-specific images across diverse categories (personalized entities, abstract concepts, domain expertise), the framework constructs structured dialogues using a synthesis model. The dialogue triplets are used to fine-tune a pretrained LVLM with weighted cross-entropy loss coefficients that balance knowledge acquisition and general capability retention.}
  \label{fig:main_figure}
\end{figure*}

\section{Method}
Given an image dataset $\mathcal{D} = \{I_1, I_2, \ldots, I_n\}$ from a specific domain, where each domain-specific knowledge is represented by only a few images (typically 3-5) without any textual labels or descriptions, our objective is to enhance the capabilities of any LVLM. We impose no restrictions on the image capture settings, allowing for diverse contextual variations in the representation of each knowledge. Our goal is to train the model to focus on these domain-specific knowledge, thereby enabling the generation of context-aware textual responses while retaining the pre-existing knowledge embedded in the pre-trained LVLM.

We begin by providing background on LVLMs (Sec. \ref{sec:3.1}), highlighting the cross-modal capabilities that enable visual understanding and identifying the key challenges in specialized knowledge acquisition. This is followed by an introduction to our SDFT technique (Sec. \ref{sec:3.2}), which employs a three-phase dialogue structure to systematically preserve foundation knowledge, establish knowledge boundaries, and inject specialized information. Finally, we propose a weighted multi-turn supervision framework (Sec. \ref{sec:3.3}) designed to balance domain-specific knowledge acquisition and general capability retention through strategically weighted loss components for each dialogue phase.
\subsection{Preliminary}
\label{sec:3.1}
Large Vision Language Models (LVLMs) are probabilistic multimodal models that integrate visual and linguistic data to perform comprehensive analysis and generation tasks. Specifically, we focus on pre-trained LVLMs designed to handle image and text pairs, where the image $\mathbf{I}$ and text prompt $\mathbf{T}$ jointly inform the model output. 

LVLMs leverage expansive datasets to learn the mapping $P(\mathbf{O} \mid \mathbf{I}, \mathbf{T})$, capturing intricate semantic correlations. These models employ deep neural architectures that merge vision encoders and text processors, optimized to support tasks such as image caption and visual question answering. A more detailed description of their mechanisms is provided in Appendix A.

\subsection{Structured Dialogue Fine-Tuning}
\label{sec:3.2}
Our primary objective is to resolve the fundamental knowledge injection dilemma in LVLMs—how to effectively incorporate domain-specific knowledge while preserving general capabilities. This challenge is particularly acute in few-shot scenarios, where conventional fine-tuning approaches lead to catastrophic forgetting \cite{tirumala2022memorizationoverfittinganalyzingtraining}. The model either becomes overly specialized, losing its foundational visual-linguistic abilities, or fails to adequately capture the nuanced aspects of the target domain knowledge. This catastrophic forgetting occurs because the transition from object-level understanding to domain-specific knowledge creates competing optimization objectives that conventional training methods cannot balance effectively

\subsubsection{Multi-turn Dialogue Architecture}
\label{sec:3.2.1}
Our SDFT framework, as illustrated in Figure \ref{fig:main_figure}, consists of three distinct dialogue turns, each serving a specific purpose in the knowledge injection process: 
\begin{enumerate}
   \item \textbf{Foundation Preservation (Turn 1)}: The first turn focuses on general image caption, reinforcing the model's pre-trained visual-linguistic alignment capabilities. For each image $I_i$, we generate a caption query $Q1(I_i)$ (e.g., "Describe this image") and its corresponding response $A1(I_i)$.
   \item \textbf{Contrastive Disambiguation (Turn 2)}: The second turn introduces a carefully designed unrelated knowledge $k_d$ unrelated to the target domain. For each image, we generate a query $Q2(I_i, k_d)$ (e.g., "How is this image related to [unrelated knowledge]?") and its corresponding negative response $A2(I_i, k_d)$ that explicitly distinguishes the image content from the unrelated knowledge.
   \item \textbf{Knowledge Specialization (Turn 3)}: The final turn directly addresses the target domain knowledge $k_t$ with a query $Q3(I_i, k_t)$ (e.g., "How is this image related to [target knowledge]?") and a detailed response $A3(I_i, k_t)$ that embeds the specialized knowledge using chain-of-thought reasoning \cite{wei2023chainofthoughtpromptingelicitsreasoning}.
\end{enumerate}

This structured dialogue design effectively mitigates catastrophic forgetting while "implanting" domain-specific knowledge into the LVLM's knowledge representation. As shown in Figure \ref{fig:main_figure}, we deliberately vary prompt structures while maintaining consistent knowledge references. For example, we alternate between questions like "How is this image related to [target knowledge]?" and "When you see this picture, do you see evidence of [target knowledge]?" This strategic variation creates robust associations between visual elements and target knowledge without causing the model to overfit to specific prompt patterns. 

Our approach creates a progressive learning path with three distinct phases. First, we anchor the model in its pre-trained distribution to maintain foundational capabilities. Next, we build clear semantic boundaries through the Contrastive Disambiguation phase, which introduces unrelated knowledge as negative examples. Finally, we inject the target domain knowledge with high-fidelity supervision. The synthesis model generates detailed responses throughout this process, explicitly connecting visual elements to their meaningful implications and creating a bridge between visual features and domain knowledge. 

This comprehensive framework effectively intertwines general vocabulary with specialized domain knowledge, leveraging the model's prior understanding while carefully expanding its semantic boundaries. By systematically progressing through these phases, our method achieves effective knowledge injection while preventing the catastrophic forgetting that typically occurs in conventional fine-tuning approaches.

\subsubsection{Dialogue Synthesis Process}
\label{sec:3.2.2}
Our dialogue synthesis process consists of two main stages that leverage both a powerful synthesis model $\mathcal{S}$ (e.g., Qwen2-VL-72B-Instruct\cite{bai2023qwen}) and the base model $\mathcal{B}$ to be fine-tuned as depicted in the left portion of Figure \ref{fig:main_figure}: :

\textbf{Stage 1: Question Generation.} 
We use the synthesis model to generate questions for each image in the following order:

\begin{align}
Q_1(I_i) &= \mathcal{S}(I_i, \text{"Generate a descriptive caption question"}) \\
Q_3(I_i, k_t) &= \mathcal{S}(I_i, \text{"Generate a question about $k_t$ } \text{"}) \\
Q_2(I_i, k_d) &= \mathcal{S}(I_i, Q_3(I_i, k_t), \text{"Replace $k_t$ with $k_d$ "})
\end{align}

Note that the prompts shown here are simplified. The complete prompting templates used in our experiments are provided in Appendix B.

\textbf{Stage 2: Response Generation.} 
For the first phase, we simply use the base model:
\begin{align}
A_1(I_i) &= \mathcal{B}(I_i, Q_1(I_i))
\end{align}

For the second phase, we employ a multi-round generation strategy to enhance reliability. The base model generates multiple responses for the same query, and we select the majority consensus: 
\begin{align}
A_2^j(I_i, k_d) &= \mathcal{B}(I_i, Q_2(I_i, k_d)) \quad \text{for } j = 1,2,...,m \\
A_2(I_i, k_d) &= \text{MajorityVote}(\{A_2^j(I_i, k_d)\}_{j=1}^m)
\end{align}
where $m=3$ in our experiments. This approach leverages repeated inference to stabilize outputs for potentially ambiguous queries.

For the third phase, we use the synthesis model to generate high-quality responses with detailed reasoning:
\begin{align}
A_3(I_i, k_t) &= \mathcal{S}(I_i, Q_3(I_i, k_t), \text{previous dialogue context})
\end{align}

This approach ensures that we maintain the base model's output distribution for general content while obtaining reliable negative responses for unrelated knowledge and high-fidelity domain information for target knowledge. As shown in the right portion of Figure \ref{fig:main_figure}, we include a quality control process that involves manual verification of the generated responses to ensure alignment with the target knowledge. 

\subsection{Weighted Multi-Turn Supervision}
\label{sec:3.3}
In standard supervised fine-tuning (SFT)\cite{ouyang2022traininglanguagemodelsfollow} for instruction tuning, the training loss is computed independently for each response in the dialogue and then summed uniformly. However, in our three-turn dialogue format, the informativeness and supervision value of each turn are inherently different. To address this, we introduce a weighted multi-turn loss formulation that explicitly balances the influence of each dialogue component, as illustrated in the upper portion of Figure \ref{fig:main_figure}. 

Let $\mathcal{L}\text{cap}$, $\mathcal{L}\text{dis}$, and $\mathcal{L}_\text{target}$ denote the cross-entropy losses computed over the model’s output distributions corresponding to the responses in the three respective phases.

We define the total training objective as:
\begin{align}
\mathcal{L}_\text{total}(\theta) = \alpha_1 \cdot \mathcal{L}_\text{cap}(\theta) + \alpha_2 \cdot \mathcal{L}_\text{dis}(\theta) + \alpha_3 \cdot \mathcal{L}_\text{target}(\theta)
\end{align}
where $\theta$ represents the model parameters, and $\alpha_1$, $\alpha_2$, and $\alpha_3$ are scalar weights that control the contribution of each turn's loss. This weighting mechanism enables fine-grained regulation of model optimization:

\begin{itemize}
    \item $\alpha_1$ emphasizes general visio-linguistic grounding via caption supervision
    \item $\alpha_2$ promotes semantic disentanglement under adversarial distraction
    \item $\alpha_3$ focuses on domain-specific knowledge injection through high-fidelity QA
\end{itemize}

We empirically set $(\alpha_1, \alpha_2, \alpha_3) = (0.2, 0.3, 0.5)$ across all tasks, which yields favorable performance trade-offs. 

\section{Experiment Settings}
\subsection{Dataset}
To rigorously evaluate the effectiveness and generalizability of our proposed method, we utilize two categories of datasets: (1) knowledge injection datasets for assessing specialized knowledge acquisition and (2) general capability evaluation datasets for measuring retention of foundational abilities.

\subsubsection{Specific Knowledge Injection Datasets}
We strategically select three knowledge injection scenarios that represent a progression of increasing abstraction and domain specificity, allowing us to evaluate our method across the full spectrum of knowledge types that may need to be injected into LVLMs:

(1) \textbf{Personalized Entities Injection Dataset.} At the most concrete level, we begin with personalized entity recognition using the dataset from \cite{nguyen2024yollavapersonalizedlanguagevision}. This represents the foundational case of knowledge injection where models must learn to identify specific instances (e.g., "my pet cat Max") while distinguishing them from general categories (e.g., "a tabby cat"). The challenge here lies in maintaining fine-grained visual discrimination without compromising general object recognition capabilities. We follow the original training and testing splits provided in the dataset.

 (2) \textbf{Abstract Concepts Injection Dataset.} Moving up the abstraction hierarchy, we next evaluate our approach on symbolic and metaphorical understanding using a multi-level visual semantics dataset \cite{yin2025insightvisioncomprehensivemultilevelchinesebased}. This middle ground of knowledge injection requires models to bridge perceptual features with abstract meanings—for instance, recognizing that an image of factory smokestacks represents "environmental pollution" rather than just describing the visible elements. This dataset tests whether our method can establish connections between concrete visual patterns and their conceptual interpretations. We specifically select subcategories with more than 60 instances, randomly sampling 10 instances per subcategory as the evaluation set.

(3) \textbf{Domain Knowledge Injection Dataset.} Finally, at the most specialized level, we construct a biomedical dataset inspired by recent domain-specific training methods \cite{cheng2025domainspecificposttrainingmultimodallarge}. The medical domain represents the most challenging knowledge injection scenario, requiring integration of specialized terminology, domain-specific reasoning patterns, and expert visual interpretation skills. For example, models must learn to identify pathological conditions in medical images and apply precise diagnostic terminology rather than relying on generic visual descriptions. Our training data is derived from two biomedical subsets $\text{PMC}^{\text{Raw}}$ \cite{zhang2025biomedclipmultimodalbiomedicalfoundation} and $\text{PMC}^{\text{Refined}}$ \cite{chen2024huatuogptvisioninjectingmedicalvisual}.

This three-tiered evaluation framework allows us to systematically analyze how our structured dialogue approach handles different knowledge types, from concrete entity recognition to abstract concept understanding to domain-specific expertise. By evaluating across this progression, we can identify whether certain knowledge categories pose unique challenges for knowledge injection and whether our method's effectiveness varies depending on the abstraction level of the target knowledge.

\subsubsection{General Capability Evaluation Datasets}
To assess retention of pre-trained capabilities and potential catastrophic forgetting, we employ three established benchmarks: POPE \cite{li2023evaluatingobjecthallucinationlarge} for measuring object hallucination tendencies, MME \cite{fu2024mmecomprehensiveevaluationbenchmark} for evaluating general multimodal reasoning abilities, and TextVQA \cite{singh2019vqamodelsread} for assessing text-in-image understanding. These benchmarks were selected to provide comprehensive coverage of diverse visual-linguistic capabilities that should be preserved during knowledge injection.

The complete dataset statistics, evaluation metrics, and data preprocessing details are provided in Appendix C.

\subsection{Data Synthesis}
We adopt Qwen2-VL-72B-Instruct as our Data Synthesizer to construct training data for all three datasets. For the domain knowledge dataset, we first extract key medical concepts (e.g., "lung cancer") from PMC-derived samples, then generate concept-specific QA pairs accordingly. The synthesizer is further applied to produce multi-turn training dialogues across all datasets. To ensure reliability, we use a three-pass generation strategy followed by majority voting, as described in Section \ref{sec:3.2.2}.
Representative examples from each dataset are provided shown in Fig \ref{fig:main_figure}.

\subsection{Models}
We conduct all experiments on two representative families of open-source vision-language models: Qwen2-VL (2B and 7B)\cite{bai2023qwen} and InternVL2 (8B) \cite{chen2024fargpt4vclosinggap}. These models are selected to ensure architectural diversity and to validate the generalizability of our approach across varying scales and design paradigms. Qwen2-VL adopts a unified vision-language architecture with strong alignment capabilities and competitive performance in general-purpose multimodal tasks, while InternVL2 features a decoupled encoder-decoder design and emphasizes fine-grained visual grounding. Evaluating our method on both families enables a comprehensive analysis of its adaptability to different model structures and pretraining strategies.

All fine-tuning experiments employed the same infrastructure as our data synthesis process. We implemented full-parameter supervised fine-tuning (SFT) rather than parameter-efficient methods, allowing comprehensive adaptation across the model architecture.

\subsection{Baseline}
We compare our approach with strong task-specific baselines across datasets. For the Personalized Entities Injection dataset, we report results from the original Yo'LLaVA paper \cite{nguyen2024yollavapersonalizedlanguagevision}, which serves as the standard benchmark for evaluating personalized visual understanding. For the Abstract Concepts Injection dataset, we implement the Yo'LLaVA approach as a comparative baseline, as no previous work has addressed this specific task. For the Domain Knowledge Injection dataset, we include two representative baselines: (1) LLaVA-Med \cite{li2023llava}, which leverages GPT-4 \cite{achiam2023gpt} to generate text-based supervision over $\text{PMC}^{\text{Raw}}$; and (2) PubMedVision \cite{chen2024huatuogptvisioninjectingmedicalvisual}, which employs GPT-4V \cite{achiam2023gpt} to construct training data based on refined PMC captions.

\section{Results}
We present experimental results across three knowledge injection scenarios: personalized entities, abstract concepts, and domain expertise.

\subsection{Personalized Entities Injection}
To evaluate our SDFT approach on personalized entity recognition, we utilized the dataset from \cite{nguyen2024yollavapersonalizedlanguagevision}, which contains multiple personalized concepts across diverse visual contexts. Our evaluation focused on two primary aspects: recognition accuracy (positive, negative, and weighted) and question-answering accuracy (visual and text-only). For each personalized concept, we trained both separate models (SDFT - Separate) and a joint model handling all concepts simultaneously (SDFT - Joint). We fine-tuned using our structured dialogue template with the weighting coefficients described in Section \ref{sec:3.3}, and compared our results against LLaVA, GPT-4V, and Yo'LLaVA baselines.

As shown in Table ~\ref{tab:personalized_entity_results}, our SDFT approach achieves competitive or superior performance compared to strong baselines in personalized entity recognition. Under separate training, SDFT attains 91.4\% positive and 94.8\% negative recognition accuracy, resulting in a weighted accuracy of 93.1\%, which outperforms Yo’LLaVA (92.4\%) and closely matches GPT-4V (92.5\%). These results confirm that SDFT achieves state-of-the-art accuracy when trained on individual entities, without requiring test-time adaptation or external retrieval modules.

Furthermore, when jointly trained on multiple entities, SDFT maintains a high weighted accuracy of 89.7\%, with only a 3.4\% drop compared to separate training. Unlike prior methods such as MyVLM and Yo’LLaVA, which require dedicated embedding training and explicit external prompts for each concept, SDFT allows multiple concepts to be injected in a unified and robust manner. Despite joint training, the model still retains high recognition accuracy for each individual concept, demonstrating strong scalability and efficient concept integration. In addition, the higher text-only QA accuracy (91.2\%) over visual QA (90.1\%) suggests that our approach effectively strengthens cross-modal alignment between visual identities and their semantic representations.

\begin{table}[t]
  \centering
  \caption{Performance comparison on personalized entity recognition and QA tasks.}
  \setlength{\tabcolsep}{5pt}
  \renewcommand{\arraystretch}{1.2}
  \begin{tabular}{l|ccc|cc}
    \toprule
    \multirow{2}{*}{\textbf{Method}} & \multicolumn{3}{c|}{\textbf{Recognition Accuracy}} & \multicolumn{2}{c}{\textbf{QA Accuracy}} \\
    \cmidrule(lr){2-4} \cmidrule(lr){5-6}
    & Pos & Neg & Weighted & Visual & Text \\
    \midrule
    LLaVA & 0.000 & 1.000 & 0.500 & 0.899 & 0.659 \\
    GPT-4V & 0.851 & 0.998 & 0.925 & 0.887 & 0.987 \\
    Yo'LLaVA & \textbf{0.949} & 0.898 & 0.924 & \textbf{0.929} & 0.883 \\
    \midrule
    SDFT (Sep.) & 0.914 & \textbf{0.948} & \textbf{0.931} & 0.901 & \textbf{0.912} \\
    SDFT (Joint) & 0.873 & 0.920 & 0.897 & 0.897 & 0.882 \\
    \bottomrule
  \end{tabular}
  \label{tab:personalized_entity_results}
\end{table}

\subsection{Abstract Concepts Understanding}
Table \ref{tab:abstract_results} presents performance results across three model architectures for abstract concept understanding tasks. The findings demonstrate significant improvements when implementing our SDFT approach across all evaluated models.

For Qwen2-VL-2B, our method achieves substantial gains in both recognition metrics and QA accuracy, with weighted recognition improving from 40.3\% to 69.3\% (+29.0\%) and QA accuracy from 42.7\% to 57.8\% (+15.1\%). Most importantly, this enhancement comes with minimal impact on general capabilities, with POPE performance even showing slight improvement (+0.6\%) and minimal degradation in TextVQA (-4.6\%).Similarly, both InternVL2-8B and Qwen2-VL-7B architectures demonstrate consistent improvements with our approach, with weighted recognition increasing by 6.9\% and 4.8\% respectively, and QA accuracy improving by 5.7\% and 3.9\%.

A critical observation across all model scales is the consistent pattern of knowledge acquisition with minimal general capability degradation. Even the most substantial decrease in general capability (TextVQA for Qwen2-VL-2B at -4.6\%) represents a favorable trade-off given the substantial gains in target concept understanding. This finding confirms that our structured dialogue approach effectively balances the injection of specialized abstract concept knowledge while preserving the models' foundational visual-linguistic capabilities.

In addition to the observed improvements, our approach markedly surpasses Yo’LLaVA in abstract concept understanding tasks. Notably, compared to Yo’LLaVA, the Qwen2-VL-2B model achieves a 22.0\% higher weighted recognition and a 16.5\% higher QA accuracy, underscoring our method's superior proficiency in tackling these complex challenges.

\begin{table*}[t]
\caption{Performance comparison on abstract concept tasks. Recognition and QA performance metrics evaluate concept understanding, while General Capability Retention measures preservation of foundational abilities across POPE, MME, and TextVQA benchmarks.}
\centering
\renewcommand{\arraystretch}{1.2}
\setlength{\tabcolsep}{5pt}
\begin{tabular}{l|l|ccc|c|ccc}
\toprule
\multirow{2}{*}{\textbf{Model}} & \multirow{2}{*}{\textbf{Method}} & \multicolumn{3}{c|}{\textbf{Recognition Performance}} & \multirow{2}{*}{\textbf{QA}} & \multicolumn{3}{c}{\textbf{General Capability Retention}} \\
\cmidrule(lr){3-5} \cmidrule(lr){7-9}
 & & \textbf{Pos} & \textbf{Neg} & \textbf{Weighted} & \textbf{Accuracy} & \textbf{POPE} & \textbf{MME} & \textbf{TextVQA} \\
\midrule
Yo’LLaVA-7B & Yo'LLaVA & 0.486 & 0.472 & 0.473 & 0.413 & -- & -- & -- \\
\midrule
\multirow{2}{*}{Qwen2-VL-2B} & Base Model & 0.386 & 0.420 & 0.403 & 0.427 & 0.872 & 0.612 & 0.680 \\
 & SDFT (Ours) & 0.529 & \textbf{0.711} & \textbf{0.693} & \textbf{0.578} & 0.878 {\footnotesize \textcolor{green}{(+0.6\%)}} & 0.608 {\footnotesize \textcolor{red}{(-0.4\%)}} & 0.649 {\footnotesize \textcolor{red}{(-4.6\%)}} \\
\midrule
\multirow{2}{*}{InternVL2-8B} & Base Model & 0.549 & 0.523 & 0.526 & 0.561 & 0.877 & 0.719 & 0.732 \\
 & SDFT (Ours) & 0.629 & \textbf{0.591} & \textbf{0.595} & \textbf{0.618} & 0.864 {\footnotesize \textcolor{red}{(-1.3\%)}} & 0.703 {\footnotesize \textcolor{red}{(-1.6\%)}} & 0.700 {\footnotesize \textcolor{red}{(-3.2\%)}} \\
\midrule
\multirow{2}{*}{Qwen2-VL-7B} & Base Model & 0.908 & 0.572 & 0.605 & 0.573 & 0.901 & 0.733 & 0.809 \\
 & SDFT (Ours) & 0.850 & \textbf{0.631} & \textbf{0.653} & \textbf{0.612} & 0.897 {\footnotesize \textcolor{red}{(-0.4\%)}} & 0.731 {\footnotesize \textcolor{red}{(-0.2\%)}} & 0.762 {\footnotesize \textcolor{red}{(-4.7\%)}} \\
\bottomrule
\end{tabular}
\label{tab:abstract_results}
\end{table*}

\subsection{Domain Expertise Integration}
\begin{table*}[t]
\caption{Biomedical domain knowledge injection performance across multiple benchmarks. Values represent accuracy (\%) on open-ended and closed-ended questions for four medical VQA datasets. General Retention measures the average accuracy across POPE, MME, and TextVQA datasets,}
\centering
\renewcommand{\arraystretch}{1.25}
\setlength{\tabcolsep}{4.5pt}
\begin{tabular}{l|l|cc|cc|cc|c|c}
\toprule
\multirow{2}{*}{\textbf{Model}} & \multirow{2}{*}{\textbf{Variant}} & \multicolumn{2}{c|}{\textbf{SLAKE}} & \multicolumn{2}{c|}{\textbf{PathVQA}} & \multicolumn{2}{c|}{\textbf{VQA-RAD}} & \textbf{PMC-VQA} & \textbf{General} \\
\cmidrule(lr){3-4} \cmidrule(lr){5-6} \cmidrule(lr){7-8} \cmidrule(lr){9-9} \cmidrule(lr){10-10}
& & OPEN & CLOSED & OPEN & CLOSED & OPEN & CLOSED & Accuracy & Retention \\
\midrule
\multirow{4}{*}{LLaVA-v1.6-8B} 
& LLaVA-Med & 0.434 & 0.623 & 0.152 & 0.477 & 0.459 & 0.563 & 0.365 & - \\
& PubMedVision & 0.500 & 0.683 & 0.170 & 0.595 & 0.425 & 0.675 & 0.404 & - \\
& AdaMLLM & \textbf{0.580} & \textbf{0.733} & \textbf{0.229} & 0.786 & 0.598 & 0.813 & 0.479 & 0.661 \\
& SDFT (Ours) & 0.570 & 0.730 & 0.225 & \textbf{0.792} & \textbf{0.602} & \textbf{0.820} & \textbf{0.485} & \textbf{0.692} \\
\midrule
\multirow{4}{*}{Qwen2-VL-2B} 
& LLaVA-Med & 0.434 & 0.555 & 0.118 & 0.381 & 0.360 & 0.511 & 0.412 & - \\
& PubMedVision & 0.500 & 0.524 & 0.178 & 0.387 & 0.370 & 0.467 & 0.458 & - \\
& AdaMLLM & \textbf{0.602} & \textbf{0.750} & 0.206 & 0.636 & \textbf{0.580} & 0.761 & 0.465 & 0.622 \\
& SDFT (Ours) & 0.550 & 0.733 & \textbf{0.229} & \textbf{0.706} & 0.571 & \textbf{0.763} & \textbf{0.467} & \textbf{0.647} \\
\bottomrule
\end{tabular}
\label{tab:biomedicine_final}
\end{table*}
Table \ref{tab:biomedicine_final} presents a comprehensive evaluation of our SDFT approach for biomedical domain knowledge injection across multiple benchmarks, comparing against established methods including LLaVA-Med, PubMedVision, and AdaMLLM.
With the LLaVA-v1.6-8B architecture, our approach demonstrates strong performance across all medical datasets. On PathVQA, SDFT achieves 79.2\% accuracy on closed-ended questions, outperforming both LLaVA-Med (47.7\%) and PubMedVision (59.5\%). Similarly, on VQA-RAD, our method reaches 82.0\% closed-question accuracy, showing substantial improvement over baseline methods. While AdaMLLM performs competitively on several metrics, our approach consistently delivers balanced performance across all benchmarks.
The most significant advantage of SDFT becomes evident in its effectiveness at mitigating catastrophic forgetting, as measured by the General Retention metric. Our method achieves 69.2\% retention with LLaVA-v1.6-8B, substantially higher than AdaMLLM's 66.1\%. This 3.1\% improvement represents a critical advancement in resolving the knowledge injection dilemma—maintaining general visual-linguistic intelligence while incorporating specialized knowledge.

\section{Ablations}

\begin{table}[t]
\centering
\renewcommand{\arraystretch}{1.2}
\setlength{\tabcolsep}{6pt}
\begin{tabular}{l|c|c}
\toprule
\textbf{Variant} & \textbf{Rec. Weighted} & \textbf{Gen. Retention} \\
\midrule
\multicolumn{3}{l}{\textbf{Response Substitution}} \\
\quad w/o substitution & 0.564 & 0.604 \\
\quad w/ substitution & \textbf{0.693} \textcolor{green}{(+12.9\%)} & \textbf{0.712} \textcolor{green}{(+10.8\%)} \\
\midrule
\multicolumn{3}{l}{\textbf{Data Synthesis}} \\
\quad Single-pass only & 0.638 & 0.621 \\
\quad Multi-round voting & \textbf{0.693} \textcolor{green}{(+5.5\%)} & \textbf{0.712} \textcolor{green}{(+9.1\%)} \\
\midrule
\multicolumn{3}{l}{\textbf{Dialogue Structure}} \\
\quad Target QA only & 0.537 & 0.589 \\
\quad Caption + Target QA & 0.591 & 0.643 \\
\quad Full (3-phase) & \textbf{0.693} \textcolor{green}{(+15.6\%)} & \textbf{0.712} \textcolor{green}{(+12.3\%)} \\
\bottomrule
\end{tabular}
\caption{Ablation studies on key components of our SDFT framework using Qwen2-VL-2B. We report weighted recognition accuracy (Rec. Weighted) and general capability retention (average of POPE, MME and TextVQA performance relative to the base model).}
\label{tab:ablation}
\end{table}

\subsection{Model Response Substitution}
To evaluate the impact of using model-generated responses during fine-tuning, we compare substituting the caption and QA responses from the fine-tuning model (our approach) versus directly using synthesizer outputs. As shown in Table~\ref{tab:ablation}, self-substitution yields substantial improvements in both weighted recognition accuracy (+12.9\%) and general capability retention (+10.8\%). This indicates that aligning the fine-tuning data with the model's own output distribution helps maintain pre-trained capabilities while improving task performance.

\subsection{Multi-Round Voting in Data Synthesis}
We compare single-pass generation with our default three-round voting strategy. Table~\ref{tab:ablation} shows that multi-round voting significantly improves both weighted recognition accuracy (+5.5\%) and general capability retention (+9.1\%). This demonstrates that enhancing supervision quality through consensus helps preserve model robustness across both specialized and general tasks.

\subsection{Dialogue Structure}
We evaluate our three-turn dialogue template against two simplified alternatives: using only the target QA, and using caption plus target QA without the contrastive turn. Table~\ref{tab:ablation} reveals that both simplifications substantially degrade performance. The full three-turn structure outperforms the caption + target QA approach by 10.2\% in weighted accuracy and 6.9\% in general capability retention. This confirms that all three turns serve crucial roles in both domain-specific learning and knowledge preservation.

\section{Analysis}
\subsection{Hidden State Representation Analysis}
\begin{figure}[t]
\centering
\includegraphics[width=1.0\linewidth]{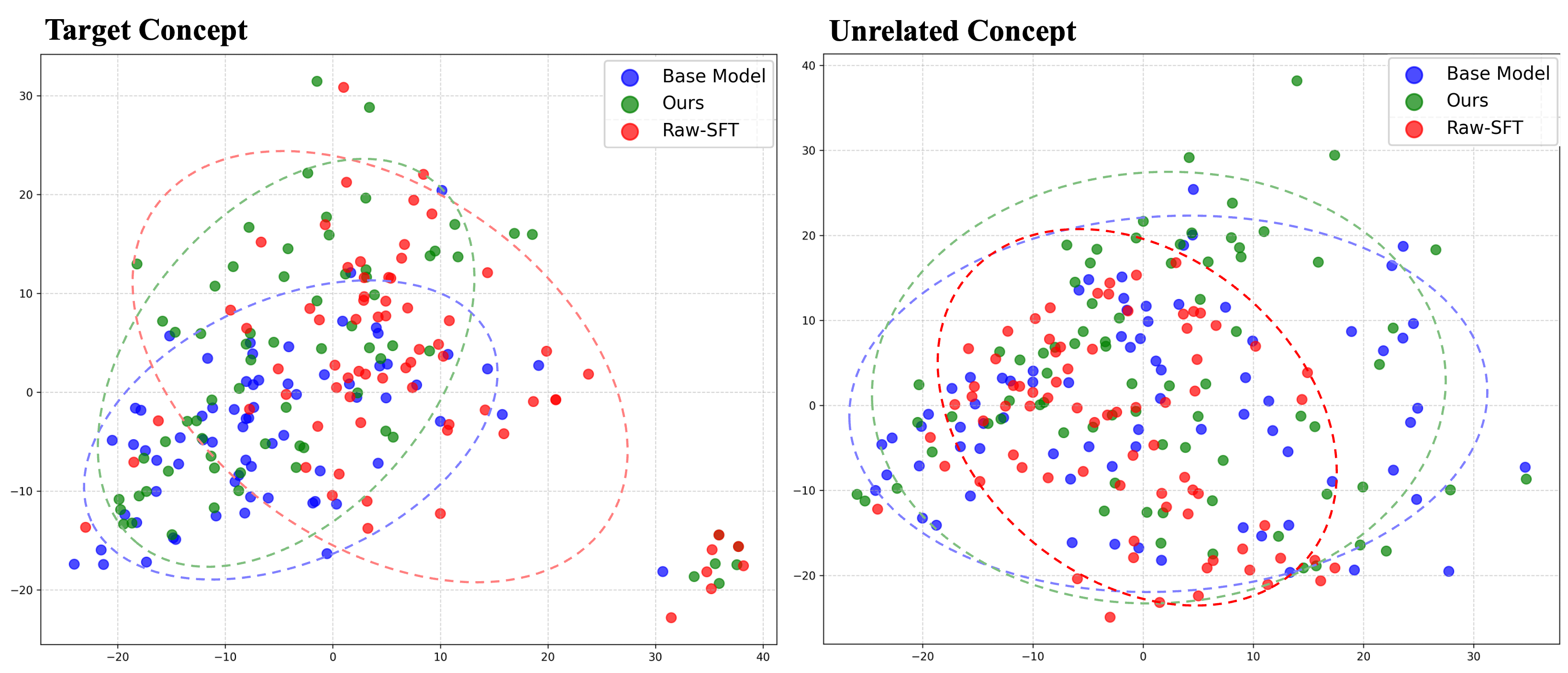}
\caption{PCA visualization of hidden states when responding to target concepts (top) and unrelated concepts (bottom). Confidence ellipses (dashed lines) indicate distribution boundaries for each approach.}
\label{fig:hidden_states}
\end{figure}

Figure \ref{fig:hidden_states} presents PCA visualizations of hidden state embeddings from three models—Base Model (blue), SDFT (green), and Raw-SFT (red)—when processing both target and unrelated concepts.
Figure \ref{fig:hidden_states} presents PCA visualizations of hidden state embeddings from three models—Base Model (blue), SDFT (green), and Raw-SFT (red)—when processing target and unrelated concepts. For target concepts (top panel), all three approaches form distinct clusters in the embedding space, with SDFT positioned intermediately between the Base Model and Raw-SFT. This strategic positioning is not merely coincidental but reflects SDFT's balanced knowledge integration approach.

The SDFT cluster demonstrates notably more cohesive organization compared to Raw-SFT's scattered distribution, indicating that our structured dialogue framework facilitates more systematic concept learning rather than arbitrary representation shifts. The confidence ellipses (dashed lines) further quantify this observation, showing that SDFT maintains a controlled deviation from the base model while Raw-SFT exhibits excessive divergence.

The unrelated concepts visualization (bottom panel) reveals an even more significant pattern: SDFT representations substantially overlap with the Base Model, while Raw-SFT deviates considerably with minimal overlap. This critical finding confirms that SDFT selectively modifies representations only for target concepts while preserving the original behavior for unrelated concepts. This selective modification capability directly addresses the catastrophic forgetting problem—SDFT effectively creates dedicated pathways for specialized knowledge while leaving general capabilities intact.

\subsection{Concept Understanding Behavior}
Qualitative analysis reveals significant differences in how models interpret abstract visual concepts. The base model consistently describes only surface-level visual elements without recognizing deeper meanings. For instance, with global warming imagery, it only identifies "smokestacks" and "smoke" without connecting these to environmental implications.

In contrast, SDFT bridges visual elements with their abstract conceptual interpretations. The model demonstrates ability to recognize that visual elements like factory emissions symbolize broader concepts such as global warming, or that raised hands in group settings represent solidarity and equality. This conceptual understanding extends beyond simple pattern recognition, as the model can articulate reasoning about how visual metaphors connect to their intended meanings . This demonstrates our dialogue structure's effectiveness in teaching conceptual understanding rather than merely improving visual feature recognition.
\subsection{Knowledge Retention Capabilities}
SDFT demonstrates superior knowledge retention while effectively integrating specialized domain knowledge. As shown in Table \ref{tab:biomedicine_final}, our approach achieves significantly better general capability retention compared to existing methods. With LLaVA-v1.6-8B, SDFT maintains 69.2\% retention, outperforming AdaMLLM's 66.1\%, while achieving comparable domain-specific performance. Similar results are observed with Qwen2-VL-2B, where SDFT maintains 64.7\% retention versus AdaMLLM's 62.2\%.

Ablation studies in Table \ref{tab:ablation} further confirm this advantage. When using only target QA pairs (Raw-SFT approach), general capability retention drops to 58.9\%, while our full SDFT framework preserves 71.2\%—a substantial 12.3\% improvement. Even when using caption and target QA without contrastive disambiguation, retention reaches only 64.3\%, highlighting each component's importance in our three-phase dialogue structure.

These results demonstrate that SDFT's structured approach creates effective knowledge boundaries that prevent interference between specialized and general capabilities, addressing the fundamental challenge of catastrophic forgetting in multimodal systems.

\section{Conclusion}
In this paper, we introduce SDFT, a novel and effective approach that resolves the catastrophic forgetting dilemma in LVLMs, enabling effective knowledge injection while preserving general capabilities. We develop a three-phase dialogue template that systematically preserves foundational abilities, establishes clear concept boundaries through contrastive disambiguation, and integrates specialized knowledge across diverse domains. Our weighted multi-turn supervision framework strategically balances knowledge acquisition with general capability retention, addressing a fundamental challenge in model adaptation. Comprehensive experiments across personalized entities, abstract concepts, and specialized domain expertise demonstrate that SDFT significantly outperforms conventional fine-tuning approaches in both specialization and capability retention. Detailed ablation studies further validate the critical contribution of each component, highlighting the effectiveness of our structured dialogue design. This versatile, model-agnostic solution offers a promising path toward building robust, domain-adapted visual AI systems without compromising their fundamental visual-linguistic intelligence. 
\section{Acknowledgments} 
This work was supported by Ant Group Research Intern Program.
\clearpage
\bibliographystyle{ACM-Reference-Format}
\bibliography{references}

\clearpage
\section{Appendix}
\section*{A. Detailed Description of LVLM Mechanisms}

Large Vision Language Models (LVLMs) represent a sophisticated class of artificial intelligence systems designed to process and integrate information across visual and linguistic modalities. These models have demonstrated remarkable capabilities in understanding complex relationships between images and text, enabling applications ranging from visual question answering to detailed image captioning and multimodal reasoning.

\subsection*{A.1 Architectural Framework}

Formally, an LVLM learns a conditional probability distribution $P(O | I, T)$, where $O$ represents the generated output, $I$ denotes the input image, and $T$ corresponds to the textual prompt. This mapping captures intricate semantic correlations between visual and textual modalities. These models typically employ deep neural architectures with four primary components:

\begin{enumerate}
    \item \textbf{Vision Encoder} $f_v: I \rightarrow \mathcal{V}$ that maps images to visual embeddings. This component is typically implemented as a convolutional neural network (CNN) or, more recently, as a vision transformer (ViT) that processes the image into a set of visual tokens or feature maps.
    
    \item \textbf{Text Processor} $f_t: T \rightarrow \mathcal{T}$ that encodes textual inputs. This component usually consists of a language model architecture such as a transformer-based encoder that converts text into dense vector representations.
    
    \item \textbf{Cross-Modal Fusion Mechanism} $f_c: (\mathcal{V}, \mathcal{T}) \rightarrow \mathcal{H}$ which integrates these representations into a unified hidden space. This fusion can take various forms, including attention-based mechanisms, concatenation followed by projection, or more complex cross-modal transformers.
    
    \item \textbf{Decoder} $f_d: \mathcal{H} \rightarrow O$ that generates the final output based on the unified multimodal representation. This component is typically a transformer-based decoder that autoregressively produces text tokens.
\end{enumerate}

\subsection*{A.2 Training Methodologies}

LVLMs are typically trained through a multi-stage process:

\begin{enumerate}
    \item \textbf{Pre-training}: Models are initially trained on large-scale image-text pairs collected from diverse sources such as web crawls, image captioning datasets, and curated multimodal collections. During this phase, the models learn general visual-linguistic associations through objectives such as masked language modeling, image-text contrastive learning, and image-conditioned text generation.
    
    \item \textbf{Instruction Tuning}: Following pre-training, models undergo alignment with human expectations through instruction-based fine-tuning. This stage involves training on multimodal instruction-response pairs that teach the model to follow user directives and generate helpful, accurate responses.
    
    \item \textbf{Preference Optimization}: Advanced LVLMs often undergo further refinement through human feedback signals, implementing techniques such as RLHF (Reinforcement Learning from Human Feedback) or DPO (Direct Preference Optimization) to align model outputs with human preferences.
\end{enumerate}

\subsection*{A.3 Challenges in Domain Adaptation}

When adapting LVLMs to specialized domains, several challenges arise:

\begin{enumerate}
    \item \textbf{Catastrophic Forgetting}: Specialized fine-tuning often causes models to lose their general capabilities as they adapt to new domains. This phenomenon occurs because updates to model parameters to accommodate new knowledge can disrupt previously learned representations.
    
    \item \textbf{Cross-Modal Alignment}: Domain-specific knowledge must be properly aligned across modalities. For instance, medical terms must be correctly associated with corresponding visual patterns in medical images.
    
    \item \textbf{Data Efficiency}: Specialized domains often lack the abundance of paired data available in general domains, necessitating efficient learning from limited examples.
    
    \item \textbf{Knowledge Boundaries}: Models must learn to distinguish when to apply specialized knowledge versus general knowledge, avoiding inappropriate application of domain-specific reasoning to general scenarios.
\end{enumerate}

The effectiveness of LVLMs is heavily dependent on the quality of the pre-training data, the alignment between visual and textual representations, and the robustness of the cross-modal fusion mechanism. Our proposed Structured Dialogue Fine-Tuning (SDFT) approach addresses these factors by systematically guiding the model through targeted dialogue interactions that preserve general capabilities while injecting specialized knowledge.
\section*{B. Prompting Templates}

Our prompting strategy differs based on the knowledge injection scenario. For domain knowledge (e.g., biomedical expertise), where numerous specialized concepts must be integrated, we employed the synthesis model to generate comprehensive dialogue templates as shown in Table~\ref{tab:domain-templates}. For personalized entities and abstract concepts, which involve fewer, well-defined concepts, we utilized structured question templates with concept substitution as detailed in Table~\ref{tab:concept-templates}.

\begin{table*}[h]
\centering
\caption{Domain Knowledge Prompting Templates Used in SDFT}
\label{tab:domain-templates}
\begin{tabular}{p{0.22\textwidth}|p{0.73\textwidth}}
\hline
\textbf{Dialogue Phase} & \textbf{Prompting Template} \\
\hline
Foundation Preservation & \textbf{User}: Describe this image in detail. \\
& \textbf{Assistant}: [Q1] \\
\hline
Contrastive Disambiguation & \textbf{User}: Modify this domain-specific question to be completely unrelated while keeping the grammatical structure. Requirements: 1. Replace key domain concepts with unrelated ones. 2. Keep the question format identical. 3. Ensure the new question cannot be answered by the original image. Original question: [Q3] \\
& \textbf{Assistant}: [Q2] \\
\hline
Knowledge Specialization & \textbf{User}: Generate a specific question that requires analyzing both the image content and knowledge of [target domain]. The question should be answerable based on the image and focus on key domain-specific elements related to [target concept]. \\
& \textbf{Assistant}: [Q3] \\
\hline
Response Generation (A3) & \textbf{User}: Here is the contextual information about the image: [domain description]. Answer the following question about this image: [Q3]. Provide a detailed response that identifies the relevant visual elements in the image, applies appropriate domain knowledge to interpret these elements, and explains the significance of these findings in relation to [target concept]. \\
& \textbf{Assistant}: [A3] \\
\hline
\end{tabular}
\end{table*}
\begin{table*}[h]
\centering
\caption{Sample Question Templates for Personalized Entities and Abstract Concepts (selected examples from our library of 200+ templates)}
\label{tab:concept-templates}
\begin{tabular}{c|p{0.85\textwidth}}
\hline
\textbf{Index} & \textbf{Question Template} \\
\hline
1 & Is there any connection between this image content and [TARGET]? \\
\hline
2 & How does this image relate to [TARGET]? \\
\hline
3 & When examining this image, can you identify [TARGET]? \\
\hline
4 & What visual elements in this image might be associated with [TARGET]? \\
\hline
5 & Does this image demonstrate or represent [TARGET] in any way? \\
\hline
6 & Can you establish any relationship between the visual content and [TARGET]? \\
\hline
7 & How might this image be interpreted in relation to [TARGET]? \\
\hline
8 & Are there visual indicators in this image that suggest a connection to [TARGET]? \\
\hline
9 & To what extent does this image convey or embody [TARGET]? \\
\hline
10 & Would you consider this image to be relevant to [TARGET]? \\
\hline
\end{tabular}
\end{table*}
\vspace{1em}
\small
\textit{Note: Table~\ref{tab:concept-templates} presents only a subset from our extensive library of prompt templates. We created a diverse set of over 200 question templates with varying phrasings to ensure robust training. During dialogue construction, we substituted the [TARGET] placeholder with either the target knowledge for Q3 or unrelated knowledge for Q2, and systematically rotated through these templates to prevent overfitting to specific question formulations. For personalized entities and abstract concepts, we created dialogue by substituting the [TARGET] placeholder with either the target knowledge (e.g., "global warming") for Q3 or unrelated knowledge (e.g., "transportation") for Q2. We systematically rotated through these templates to ensure robustness against specific phrasings.}
\normalsize

\section*{C. Experimental Details}
\subsection*{C.1 Dataset Statistics}

Table~\ref{tab:domain-templates} summarizes the statistics of the datasets used in our experiments on abstract concepts.

\subsection*{C.2 Evaluation Metrics}

To comprehensively assess both knowledge injection effectiveness and general capability retention, we employed the following evaluation metrics:

\begin{itemize}
    \item \textbf{Recognition Accuracy}: Measures the model's ability to correctly identify the presence or absence of specific knowledge concepts in images.
    \begin{itemize}
        \item \textit{Positive Recognition Accuracy}: The proportion of correctly identified instances where the target knowledge concept is present in the image.
        \item \textit{Negative Recognition Accuracy}: The proportion of correctly identified instances where the target knowledge concept is not applicable to the image.
        \item \textit{Weighted Recognition Accuracy}: A balanced measure calculated as the weighted average of positive and negative recognition accuracies, accounting for potential class imbalance in the evaluation set.
    \end{itemize}
    
    \item \textbf{QA Accuracy}: Evaluates the model's capacity to correctly answer questions about specific knowledge concepts in relation to visual content. This metric assesses not only concept recognition but also the depth of understanding and ability to articulate concept-specific reasoning.
    
    \item \textbf{General Capability Retention}: Quantifies the preservation of pre-trained capabilities through performance on established benchmarks:
    \begin{itemize}
        \item \textit{POPE} \cite{li2023evaluatingobjecthallucinationlarge}: Measures object hallucination tendencies, calculated as the average of precision, recall, and F1 scores across multiple object categories. This metric reveals whether the model maintains accurate object recognition capabilities without fabricating non-existent objects.
        
        \item \textit{MME} \cite{fu2024mmecomprehensiveevaluationbenchmark}: Evaluates general multimodal reasoning abilities across perception, knowledge, and reasoning dimensions. We report the average score across all MME subcategories to provide a comprehensive assessment of general multimodal intelligence.
        
        \item \textit{TextVQA} \cite{singh2019vqamodelsread}: Assesses text-in-image understanding capabilities, measuring the model's ability to read and reason about textual elements within images.
    \end{itemize}
\end{itemize}

For domain-specific evaluations, we also employed specialized metrics:

\begin{itemize}
    \item \textbf{Open-ended Question Accuracy}: For medical VQA datasets, we evaluate the semantic correctness of answers to open-ended questions using BERTScore with a threshold of 0.85, allowing for variations in medical terminology while maintaining semantic equivalence.
    
    \item \textbf{Closed-ended Question Accuracy}: For questions with definitive answers (e.g., yes/no or multiple choice), we calculate exact match accuracy, with partial credit assigned for answers that contain the correct option but include additional information.
\end{itemize}

Relative performance changes are reported against the base model to quantify both knowledge acquisition (improvements in recognition and QA metrics) and potential capability degradation (decreases in general capability metrics).
\end{document}